# Alan Turing and the "Hard" and "Easy" Problem of Cognition: Doing and Feeling

Stevan Harnad
Canada Research Chair in Cognitive Sciences
Département de psychologie
Université du Québec à Montréal
&
University of Southampton, UK

**ABSTRACT:** *The "easy" problem of cognitive science is explaining how and why we can do what we can do. The "hard" problem is explaining how and why we feel. Turing's methodology for cognitive science (the Turing Test) is based on doing: Design a model that can do anything a human can do, indistinguishably from a human, to a human, and you have explained cognition. Searle has shown that the successful model cannot be solely computational. Sensory-motor robotic capacities are necessary to ground some, at least, of the model's words, in what the robot can do with the things in the world that the words are about. But even grounding is not enough to guarantee that -- nor to explain how and why -- the model feels (if it does). That problem is much harder to solve (and perhaps insoluble).*

Alan Turing made countless invaluable and eternal contributions to knowledge -- the computer, computation, limits of provability, neural nets, the Turing test, breaking the Enigma code that helped save the world from Nazi tyranny -- before unspeakable injustice and ingratitude ended his short life.

I want to enlarge on just one thread in all he has done: The Turing Test set the agenda for what later came to be called "cognitive science" -- the reverse-engineering of the capacity of humans (and other animals) to think.

What is thinking? It is not something we can observe. It goes on in our heads. We do it, but we don't know how we do it. We are waiting for cognitive science to explain to us how we -- or rather our brains -- do it.

What we can observe is what we do, and what we are capable of doing. Turing's contribution was to make it quite explicit that our goal should be to explain how we can do what we can do by designing a model that can do what we can do, and can do it so well that we cannot tell the model apart from one of us, based only on what it does and can do. The causal mechanism that generates the model's doing-capacity will be the explanation of thinking, intelligence, understanding, knowledge -- all just examples of, or synonyms for: cognition.

Turing actually formulated (what eventually came to be called) the Turing Test (TT) somewhat differently. He called it the "Imitation Game," and in order to rule out any bias that might influence our judgment because of the way the TT candidate looked -- rather than just what it could do -- the test was to be purely verbal, via the exchange of written messages, with the candidate out of sight. Today we would say that the test had to be conducted via email: Design a system that can communicate by email, as a pen-pal, indistinguishably from a human, to a human, and you have explained cognition.

Questions arise: (1) Communicate about what? (2) how long? (3) with how many humans?

The answers, of course, are: (1) Communicate about anything that any human can communicate about verbally via email, (2) for a lifetime, and (3) with as many people as any human is able to communicate with.

This is a tall order, and it still leaves open the fourth question: (4) How? The answer, of course, will be to design the winning model, and cognitive science is nowhere near being able to do that, but there is a sub-question about what kind of system the winning model will be.

Many people have assumed that Turing had meant and expected the TT-passer to be a purely computational system. Computation, as Turing taught us, is the manipulation of symbols (e.g., 0's and 1's, but they could also be words) on the basis of purely formal rules that operate only on the shapes of the symbols, not their meaning (i.e., syntax, not semantics).

An example of such a formal, shape-based rule is: IF YOU READ "1 + 1  =" THEN WRITE "2".

You don't need to know what "1" or "+" means in order to follow that rule. You just need to know what to do with the shapes.

That's computation. And that's basically what a "Turing Machine" (the abstract precursor of the computer) does.

But did Turing really mean that he thought cognition would turn out to be just computation? The "computationalists" among contemporary cognitive scientists think cognition is just computation, but I don't think Turing did. The Turing Test as he described it was just an email pen-pal test: only symbols in and symbols out. That does leave the possibility that the only thing needed in between, to successfully pass the test, is symbol-manipulation (computation).

But the philosopher, John Searle showed, with his famous "Chinese Room" thought-experiment, that this cannot be true: cognition cannot be just computation. For if just a computer program were enough to pass the Turing Test, Searle himself could show that that would not generate understanding in the system that was passing the Turing Test:

Searle asks us to suppose that the Turing Test (TT) is conducted in Chinese (Chinese email, with real Chinese pen-pals). Now since computation is just rules for manipulating symbols based on their shapes, not their meanings, Searle himself could memorize and execute that same TT-passing computer program, yet he would not be understanding Chinese. But then neither would the computer that was executing the TT-passing program. Hence cognition is not just computation.

What is missing to make symbols meaningful, the way words and thoughts are meaningful to us? I've dubbed this the "symbol grounding problem": Consider a Chinese-Chinese dictionary. It defines all the words in Chinese. But if you don't already know at least the meaning of some Chinese words, the definitions of the meaningless symbols only lead to more meaningless symbols, not to meaning. Some of the symbols, at least, have to be "grounded" in what the symbols denote directly, rather than just via meaningless, formal verbal definitions.

Consider the symbol string "'zebra' = 'horse' + 'stripes'." To be able to understand that definition, you have to already know what "horse" and "stripes" mean. And that can't go on via just definitions all the way down ("stripes" = "horizontal" + "lines," etc.). Some words have to be grounded directly in our capacity to recognize, categorize, manipulate, name and describe the things in the world that the words denote. This goes beyond mere computation, which is just formal symbol manipulation, to sensorimotor dynamics, in other words, not just verbal capacity but robotic capacity.

So I do not believe that Turing was a computationalist: he did not think that thinking was just computation. He was perfectly aware of the possibility that in order to be able to pass the verbal TT (only symbols in and symbols out) the candidate system would have to be a sensorimotor robot, capable of doing a lot more than the verbal TT tests directly, and drawing on those dynamic capacities in order to successful pass the verbal TT.

But although Turing was not a computationalist about cognition, he was nevertheless a computationalist in the more general sense that he believed that just about any physical, dynamical structure or process (including planetary motion, chemical reactions, and robotic sensorimotor dynamics) could be simulated and approximated by computation as closely as we like. This is called the physical version of the "Church-Turing" Thesis (CT). (The mathematical version of CT is the thesis that

Turing's formal definition of computation -- the Turing Machine -- can do anything and everything that mathematicians do when they "compute" something.)

The physical CT does not imply, however, that everything in the physical world is just computation, because everyone knows that a computer simulation of (say) a plane, is not a plane, flying (even if it can simulate flying well enough to help test and design plane prototypes computationally, without having to build and test them physically, and even if the computation can generate a virtual reality simulation that the human senses cannot distinguish from the real thing -- till they take off their goggles and gloves).

So Searle is simply pointing out that the same is true of computational simulations of verbal cognition: If they can be done purely computationally, that does not mean that the computations are cognizing.

Computations cognizing? What on earth does that mean? Well, to answer that question, we have to turn to another philosopher: Descartes. How does Searle know that he is not understanding Chinese when he is passing the Chinese TT by memorizing and executing the TT-passing computer program? It is because it feels like something to understand Chinese. And the only one who knows for sure whether that feeling (or any feeling at all) is going on is the cognizer -- who is in this case Searle himself.

The contribution of Descartes' celebrated "Cogito" is that I can be absolutely certain that I am cognizing when I am cognizing. I can doubt anything else, including what my cognizing seems to be telling me about the world, but I can't doubt that I'm cognizing when I'm cognizing. That would be like doubting I'm feeling a toothache when I am feeling a toothache: I can doubt whether the pain is coming from my tooth -- it might be referred pain from my jaw -- I may not even have a tooth, or a mouth, or a body; there may be no outside world, nor any yesterday or tomorrow. But I cannot doubt that what it feels like right now is what it feels like right now.

Well Searle is not feeling the understanding of Chinese when he passes the Chinese TT. He can distinguish real understanding (as he understands English) from just going through the motions: just doing the doing.

But where does this leave Turing's test, then, which is based purely on doings and doing-capacity, indistinguishable from the doing capacity of real, cognizing human beings?

Turing was perfectly aware that generating the capacity to do does not necessarily generate the capacity to feel. He merely pointed out that explaining doing power was the best we could ever expect to do, scientifically, if we wished to explain cognition. The successful TT-passing model may not turn out to be purely computational; it may be both computational and dynamic; but it is still only generating and explaining our doing capacity. It may or may not feel.

Explaining how and why we can do what we can do has come to be called the "easy" problem of cognitive science (though it is hardly that easy, since we are nowhere near solving it). The "hard" problem is explaining how and why we feel -- the problem of consciousness -- and of course we are even further from solving that one.

In commemoration of Turing's 2012 centenary the Cognitive Sciences Institute of Universite du Quebec a Montreal is hosting a 10-day Summer Institute on the [Evolution and Function of Consciousness](#) (plus a 3-day practical workshop on measuring consciousness) from June 29 to July 12 in Montreal. Over 50 computer scientists, roboticists, neuroscientists, biologists, psychologists and philosophers (including [John Searle](#), [Dan Dennett](#), [Antonio Damasio](#), [Joseph Ledoux](#) and [Simon Baron-Cohen](#)) will present the current state of the art in the attempt to give a causal explanation of how and why we feel rather than just do. For those who cannot attend in person, the videos of most of the talks will be available on the web as of the day after each presentation.

Harnad, S. (1992) [The Turing Test Is Not A Trick: Turing Indistinguishability Is A Scientific Criterion](#). *SIGART Bulletin* 3(4) (October 1992) pp. 9 - 10.